\documentclass{ieeeconf}

{}
{}
{}

\usepackage{times}
\usepackage{epsfig}
\usepackage{graphicx}
\usepackage{amsmath}
\usepackage{amssymb}
\usepackage{subfigure}
\usepackage{setspace}
\usepackage{mathtools} 
\usepackage{algorithmicx}
\usepackage{multirow}
\usepackage[ruled]{algorithm}
\usepackage{color, colortbl}
\definecolor{LightGray}{rgb}{0.9,0.9,0.9}
\usepackage{url}

\newcommand{\etal}{\mbox{\emph{et al.}}}

\IEEEoverridecommandlockouts                              % This command is only needed if 
\title{\LARGE \bf
	Exploiting Feature Confidence for Forward Motion Estimation
}

\author{Chang-Ryeol Lee and Kuk-Jin Yoon % <-this % stops a space
	\thanks{C. Lee, and K. Yoon are with the School of Information and Communications
		at the Gwang-ju Institute of Science and Technology, Gwang-ju, Korea
		{\tt\small\{crlee, kjyoon\}@gist.ac.kr}} % <-this % stops a space
}

\begin{document}

\maketitle
\thispagestyle{empty}
\pagestyle{empty}

%\supertitle{Regular Papers}

%\title{Visual-Inertial Navigation with Feature Confidence Analysis}
%
%\author{\au{Chang-Ryeol Lee}, \au{Kuk-Jin Yoon$^{\corr}$}}
%
%\address{{School of Electrical Engineering and Computer Science, Gwangju Institute of Science and Technology (GIST), 123 Cheomdangwagi-ro, Buk-gu, Gwangju, Korea}\\
%%\add{4}{Current affiliation: Fourth Department, Fourth University, Address, Country Name}
%\email{kjyoon@gist.ac.kr}}

\begin{abstract}
%{Odometry, which estimates 6-DOF ego-motion, is a crucial technology for mobile applications.}
%
Visual-Inertial Odometry (VIO) utilizes an Inertial Measurement Unit {(IMU)} to overcome the limitations of Visual Odometry (VO).
However, the VIO for vehicles in large-scale outdoor environments still has some difficulties in estimating forward motion with distant features. % {along the motion axis.}
To solve these difficulties, we propose a robust VIO method based on the analysis of feature confidence in forward motion estimation using {an IMU.} 
%inertial measurement units. 
%
We first formulate the VIO problem by using effective trifocal tensor geometry. 
Then, we {infer the feature confidence by using the motion information obtained from {an IMU}  and incorporate the confidence into the Bayesian estimation framework.}
Experimental results on the public KITTI dataset show that the proposed {VIO} outperforms the baseline VIO,
and it also demonstrates the effectiveness of the proposed feature confidence analysis and confidence-incorporated ego-motion estimation framework. 
\end{abstract}

%\maketitle

\section{Introduction}\label{sec:intro}

% Introduction of visual odometry and visual-inertial odometry
Odometry, which estimates 6-DOF ego-motion, is a crucial technology for mobile applications{.
	I}n robotics and computer vision community, Visual Odometry (VO) using cameras has been extensively studied for robot navigation \cite{Moravec:IJCAI:1979} and autonomous driving \cite{Geiger:CVPR:2012} for decades.   
Practically, VO has great advantages in GPS-denied environments such as urban, military, underwater, and indoor areas, and provides less drifted results compared to Wheel Odometry (WO) and Inertial Odometry (IO).
However, it also has some limitations: it cannot estimate the absolute scale of ego-motion and its performance highly depends on the scene and/or surrounding environments.
%

%% The utilization of IMU in computer vision community
%{These limitations can be overcome to some extent by exploiting more information from other sensors. For example, an} inertial sensor, which measures acceleration and angular velocity, is practically helpful for various computer vision applications {because it} provides reliable motion measurements {in real-time, regardless of weather and illumination conditions} without any processing. 
%%
%{For example,} Hwangbo {\textit{et al.} \cite{Hwangbo:IROS:2009} used motion information from {an inertial measurement unit (IMU)} for robust feature tracking in the circumstance {where} motion {changes} abruptly.
%%
%Joshi \textit{et al.} \cite{Joshi:SIG:2010} exploited IMU measurements for accurate image de-blurring.
%%
%Kurz and BenHimane \cite{Kurz:CVPR:2011} proposed feature descriptors based on gravity information from {an} IMU.
%%

For these reasons, Inertial Measurement Units (IMUs) have been utilized for VO, named Visual-Inertial Odometry (VIO), and led outstanding advances in ego-motion estimation \cite{Kneip:BMVC:2011}.
The VIO exploits the inertial measurements as well as a camera to get the scale information and accurate rotation estimates.
It shows more robust performance to variations of scene conditions (\textit{e.g.} illumination, weather, texture, etc.) than VO thanks to the complementary properties of heterogeneous sensors \cite{Hu:ICRA:2014}. 

\begin{figure}[t]
	\centering
	\includegraphics[width=0.99\linewidth]{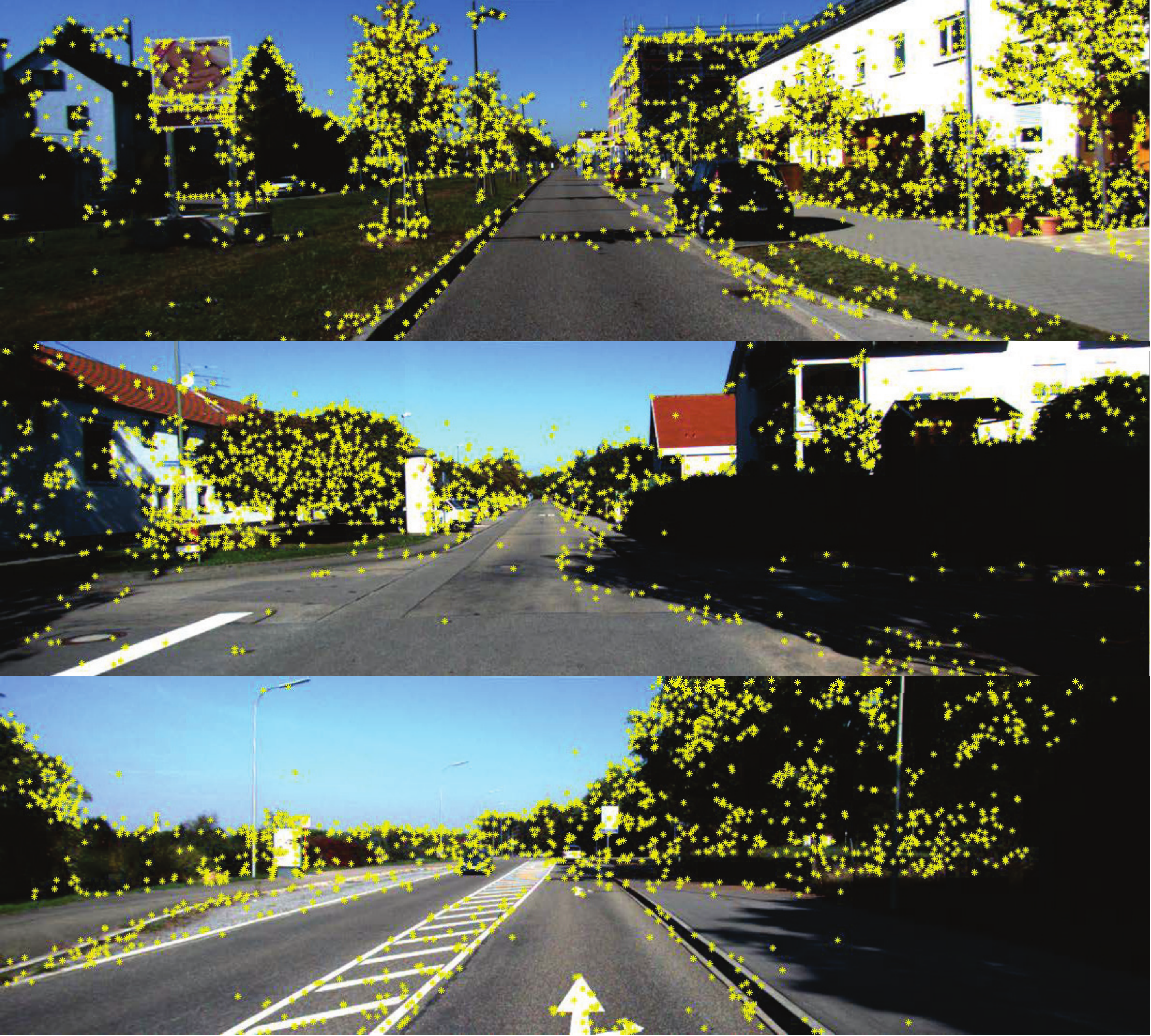}
	\caption{Some outdoor images from the KITTI vehicle sequences and extracted local features. Many feature points are located near the motion axis and distant in the scene. { Such distant feature points near the motion axis have large uncertainty for forward motion estimation.}}
	\label{fig:example_vehicle_image}
	\vspace{-5mm}
\end{figure}

% Problem
However, {despite the aid of {an} IMU, these vision-based odometry methods}  \emph{for vehicles in large-scale outdoor environments} still {have} some difficulties in estimating forward motion with distant features, 
%However, the VIO \emph{for vehicles in large-scale outdoor environments} {still has some difficulties in estimating translational motion with distant features.} 
%
where accurate estimation of the forward motion is a crucial issue {because vehicles (\textit{i.e.} cameras) move mostly forward.} 
When they move forward and only distant features are available in the scene, it causes the lack of observation for translational motion estimation.
% 1) estimating forward/backward motion and
For example, feature points located along the optical axis do not have any displacement{s} in the images when a camera {moves along the optical axis.} 
In this sense, \emph{features near the motion axis are not informative for forward motion estimation.}
%
% Probelem2: large-depth-feature
Furthermore, distant features have small displacements {under translational camera motion.}
%
%{It means that the signal-to-noise ratio (SNR) of the distant feature points is larger than that of close features.}
%
%while features with small displacements between frames are easy to track for ego-motion estimation, they are likely to have large depth as shown in Fig. \ref{fig:example_vehicle_image} in the case of translational camera motion.
%
{Actually, it is a well-known fact} that {\emph{distant features are not informative for translation motion estimation}} \cite{Kaess:ICRA:2009}.
Consequently, {\emph{ \textbf{distant features along the motion direction}}} have vast uncertainty {for forward} motion estimation due to the aforementioned reasons.
Ironically, however, such uncertain features are easy to track in the image sequence because of the projectivity of an image and their small displacement under translation as shown in Fig. \ref{fig:example_vehicle_image}.

Yang \etal \cite{Yang:CVPRW:2010} analyzed the uncertainty of a 3D feature point, $U_p$, during forward motion for scene reconstruction from motion as
\begin{equation} \label{eq:uncertainty_feature}
	U_p = \frac{\sigma}{t_3R}\sqrt{\frac{N}{X^2+Y^2}}, \ N = f(X,Y,Z),
\end{equation}
where $\sigma$ is the standard deviation of measurement noise, $R$ is the distance of a 3D feature point, $t_3$ is the forward translation of a camera, and $(X,Y,Z)$ is a 3D location of a feature point. 
Here, a composition function $f$ is dominated by $Z^8$, and this indicates that the uncertainty of the measurement increases as the depth of a feature point increases. 
In addition, we can see that feature points around the motion axis have large uncertainty because $X,Y,R$ are placed in the denominator.

\begin{figure}[t]
	\centering
	\includegraphics[width=0.99\linewidth]{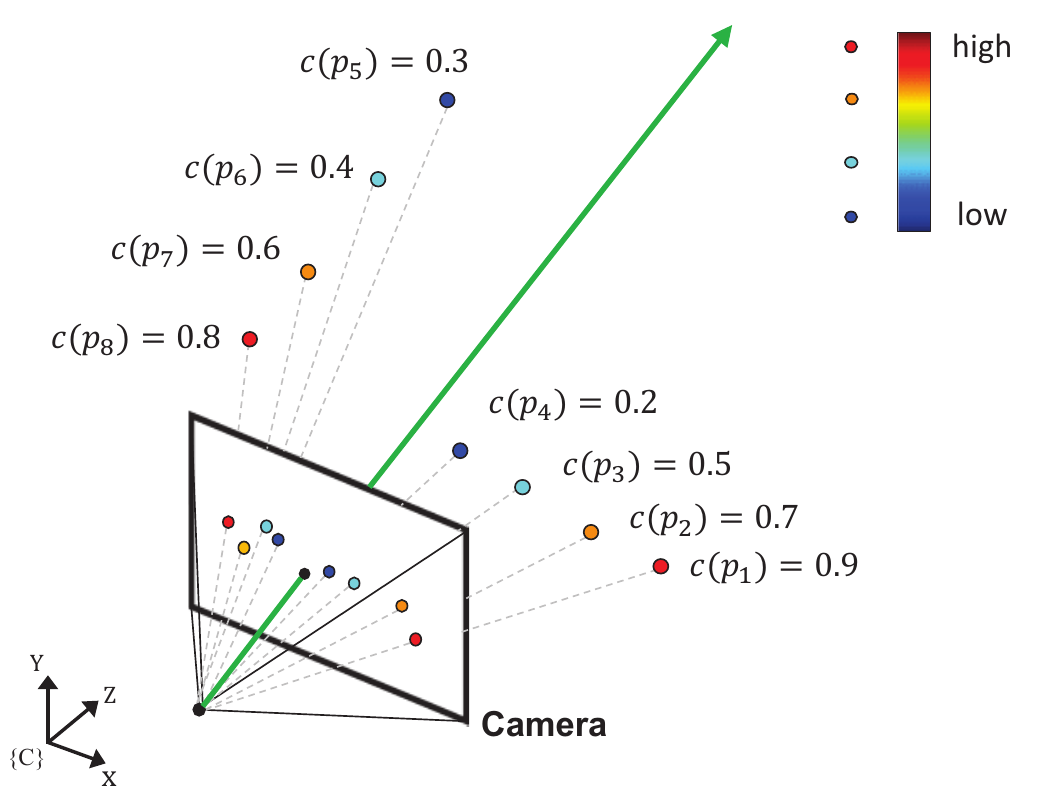}
	\caption{Concept of a confidence map during forward motion. {As the depth (Z-axis) of the feature increases, the confidence of the feature decreases. In addition, as the distance of the feature from the motion direction increases, the confidence of the feature increases. }}
	%	{As the angle between the motion direction vector and the feature position vector increases, the confidence of a feature in motion estimation decreases.}}
	\label{fig:concept of confidence}
\end{figure}

% Key idea of proposed method
Our work was inspired by this simple analysis on {the} uncertainty of measurements.
In this paper, we define the confidence of a feature point in motion estimation as the opposite concept of uncertainty due to their inverse correlation.
{Fig. \ref{fig:concept of confidence} shows the concept of the feature confidence when {a} camera moves forward}. % by (\ref{eq:uncertainty_feature}) on {the} uncertainty of feature{s}.}
{We presents} a robust VIO method in a moving platform by analyzing the confidence of features {during forward motion}.
The cue for the confidence is the motion direction obtained from inertial measurements, and it is incorporated into measurement noise covariance in {the Bayesian estimation framework.}
Our confidence analysis can be easily utilized for {existing vision-based odometry frameworks for vehicles with inertial measurements or pre-estimated motion from images.}

The main contributions of the paper can be summarized as follows:

% CR: contribution 

$\bullet$ Analysis of feature confidence {in forward motion estimation} based on inertial measurements, and  %with {the} simple approximation on Yang's analysis \cite{Yang:CVPRW:2010}.\\

%$\bullet$ Confidence-incorporated RANSAC process based on inertial measurements, and

$\bullet$ {Confidence-incorporated {Bayesian} ego-motion estimation framework.}

% composition of paper
This paper is organized as follows.
We review related works in Section \ref{sec:related_works} {and describe the system configuration and notations in} Section \ref{sec:notation_configuration}.
Then, we briefly summarize {the baseline VIO framework} \cite{Hu:ICRA:2014} in Section \ref{sec:trifocal_tensor_based_VIO}. 
We describe how the confidence of feature{s} can be inferred and {incorporated into our framework} in Section \ref{sec:confidence_analysis}, and illustrate experimental results in Section \ref{sec:experimental_results}.
Finally, we discuss the limitation and future works and conclude the paper in Section \ref{sec:conclusion}.

%========================================================

\section{Related Works}\label{sec:related_works}

We review VIO methods and feature handing methods for vision-based odometry.
{\flushleft{\textbf{Visual-Inertial Odometry}}}. 
{In the early stage}, inertial measurements {were} used for structure-from-motion applications \cite{Qian:ICIP:2002,Gemeiner:ICRA:2005,Jones:ICCV:2007}. 
Those methods simply integrated inertial measurement data into {the} Bayesian {estimation} framework to predict ego-motion.
Then, the fusion of visual and inertial measurement data {has been} theoretically {validated} and various state estimation techniques {have been also} applied \cite{Huang:IJRR:2010,Kneip:ICRA:2011,DongSi:ICRA:2011,Jones:IJRR:2011}.
{These} methods usually require heavy computational power because of {the} estimation of 3D landmarks. 
{As a result,} a real-time issue on VIO {has become important}, and {many} efficient algorithms {have been} studied \cite{Weiss:ICRA:2011,Lupton:TRO:2012}. 

Nowadays, the inertial measurements are more effectively used in various ways for many applications.
{For example,} the inertial data were used to solve {an} inverse depth problem in SLAM \cite{Pinies:ICRA:2007}. 
Also, researches to estimate {the} scale of camera motion by using inertial data were studied \cite{Nutzi:JIRS:2011,Kneip:ICRA:2011,Martinelli:TRO:2012}. % 
{In some works, the calibration of IMUs and between an IMU and a camera, which directly affects on estimation performance, {was} estimated together with the motion} \cite{Jones:ICCV:2007,DongSi:IROS:2012,Li:ICRA:2013}.
Feature characteristics {was} often exploited as constraints for estimation \cite{Williams:ICRA:2011,Kottas:ICRA:2013},
and there {was} some efforts to resolve issues arisen in {smart-phones} \cite{Li:IROS:2012,Li:ICRA:2013}.
Recently, sliding-window approaches {have been proposed for VIO} \cite{Mourikis:ICRA:2007,DongSi:ICRA:2012,Hu:ICRA:2014},   
{and} optimization-based VIO methods {have been also studied} \cite{Li:RSS:2012,Leutenegger:IJRR:2015}.
{\flushleft{\textbf{Feature Handling for Odometry}}}.
Kaess \etal \cite{Kaess:ICRA:2009} proposed {the} separated estimation of rotation and translation to handle degenerate {cases where} features are not distributed evenly. 
They used depth cue based on stereo {cameras}.
Badino \etal \cite{Badino:ICCV:2013} pointed out that handling the noise of feature {measurement} is an important issue {and tried to suppress} the noises of features by integrating features {in} multiple frames. 
{On the other hand}, Song \etal \cite{Song:CVPR:2014} {used a} learning technique to resolve the different uncertainties of multiple cues.
%
% CR: CVPR 2016, good feature to track for visual slam
{Recently,} Zhang and Vela proposed {to} select important {features} for {ego-motion estimation} \cite{Zhang:CVPR:2015}.
{This is the most similar work to ours.}
{However, unlike Zhang's algorithm,}  we exploit simple confidence analysis based on inertial measurements {instead of complex observability measure.}

%========================================================

\section{System Configuration and Notations}\label{sec:notation_configuration}

\begin{figure}[tb]	
	\centering
	\includegraphics[width=0.7\linewidth]{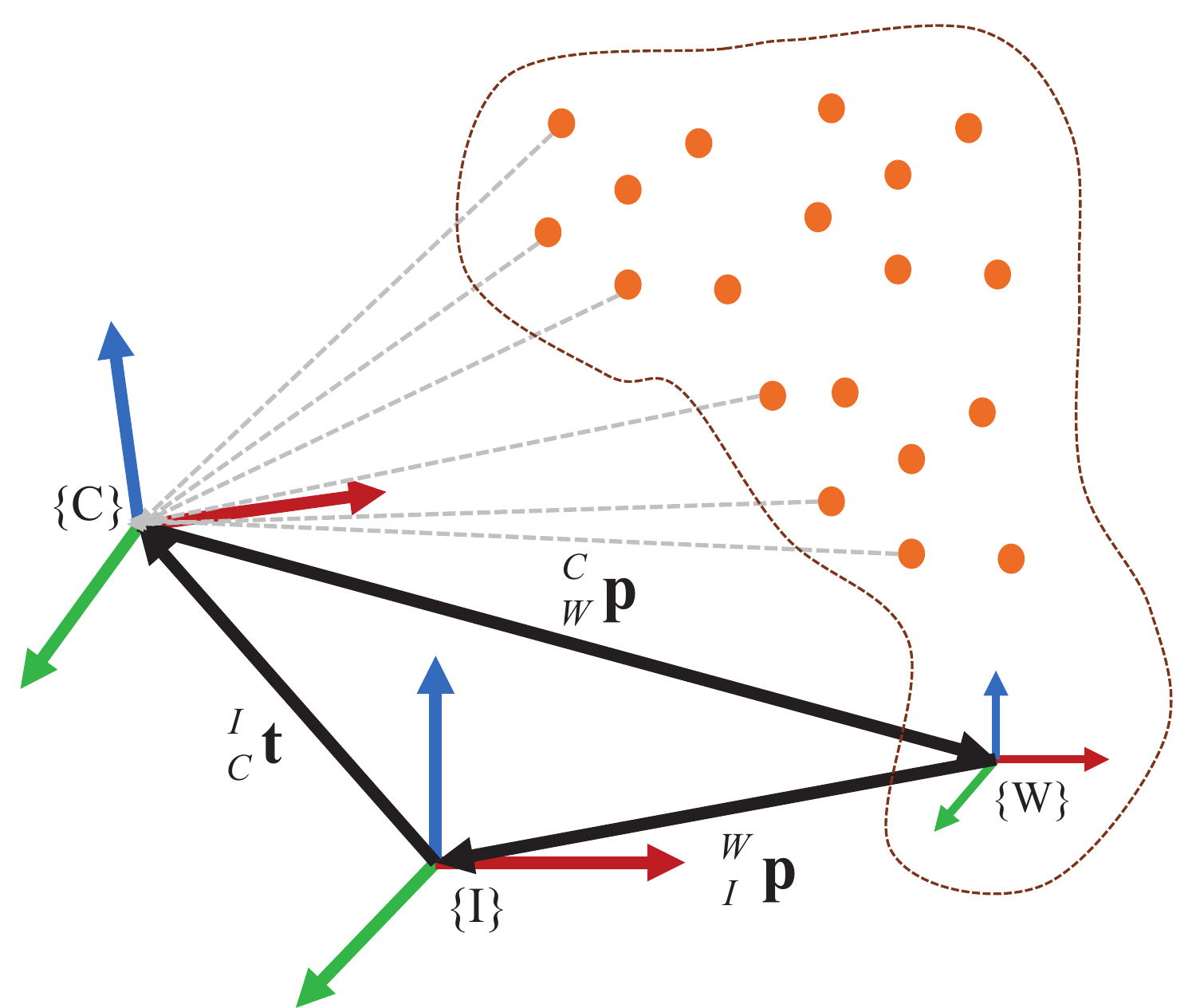}
	\caption{Relation between 3D feature points in world coordinates and rigidly connected IMU-camera coordinates. {Orange dots indicate feature points in world coordinates. Dashed gray arrows represent the projection of the features to camera coordinates.}}
	\label{fig:IMU_cam_configuration}
\end{figure}

Before presenting the problem formulation and {the} proposed algorithm, we {describe} the system configuration and some notations to make our formulation clearer. 
Our system {consists} of a single IMU and a single camera.
They are rigidly connected {and, therefore,} {the transformation between their coordinates can be described by} 3-DOF rotation and 3-DOF translation.
Fig. \ref{fig:IMU_cam_configuration} shows the relation between 3D feature points in {the} world coordinate and rigidly connected IMU-camera coordinates.
$\{I\}$ denotes the IMU coordinate, and $\{W\}$ represents the world coordinate. 
Special subscripts (\textit{i.e.}, $^W_I$ and $^I_C$) explain the {reference} coordinate explicitly. 
For example, when $\mathbf{p}$ denotes the 3D position, $^W_I\mathbf{p}$ represents the IMU 3D position with respect to the world coordinate $\{W\}$; here, the subscript $I$ represents an IMU. 
When $\mathbf{t}$ denotes 3D translation, $^I_C\mathbf{t}$ represents camera translation with respect to the IMU coordinate $\{I\}$; here, the subscript $C$ represents a camera.

%========================================================

\section{Visual Inertial Odometry}\label{sec:trifocal_tensor_based_VIO}

% CR: revise 
In the ego-motion estimation problem, 3D landmarks are commonly estimated together with the 6D motion vector.
This approach requires a large amount of computational power proportional to the number of landmarks in general.
%
%Our algorithm is based on trifocal tensor between consecutive frames which is effective on monocular camera and a IMU.
{For efficient and accurate ego-motion estimation,} our work is based on {the} sliding window strategy that uses consecutive frames without estimating any 3D landmarks.
We exploit {trifocal-tensor-based VIO \cite{Hu:ICRA:2014} as a baseline odometry algorithm {with little modifications.} 
% It is popular Kalman filter based state estimation technique which use three consecutive frame \cite{Hu:ICRA:2014}. 
% 
{In this section, we briefly summarize the baseline algorithm.}

{Basically,} we estimate the state $\mathbf{x}$ in  (\ref{equation:state}) by using the measurement $\mathbf{z}$ in  (\ref{equation:modified_VIO:measurement_model_3}) 
with the Unscented Kalman Filter (UKF) because both the state transition and {measurement} models are nonlinear.
The state comprises 3D poses of an IMU in three consecutive frames for trifocal tensor geometry {and biases of inertial measurements}:
\begin{equation}
	\label{equation:state}
	\textbf{x} =
	\left[  \ {^W_I\textbf{p}_{1}^\top} \ {^W_I\textbf{v}^\top} \  
	{^W_I\textbf{q}_{1}^\top}  \ \textbf{b}_{a}^\top \ \textbf{b}_{g}^\top 
	\ {^W_I\textbf{p}_{2}^\top} \  {^W_I\textbf{q}_{2}^\top}       
	\ {^W_I\textbf{p}_{3}^\top} \  {^W_I\textbf{q}_{3}^\top} \  
	\right]^\top,
\end{equation}
where $^W_I\textbf{q}_i \in \mathfrak{R}^4$ is a unit quaternion representing the orientation, $^W_I\textbf{p}_i \in \mathfrak{R}^3$ is the position,  {and the subscript $i$ ({$i \in \{1,2,3\}$}) denotes the index of {the} poses in order of time ( \textit{i.e.}, $^W_I\textbf{q}_{1}$ and $^W_I\textbf{p}_{1}$ are the foremost pose).} 
$^W_I\textbf{v} \in \mathfrak{R}^3$ is the velocity, {and $\textbf{b}_{a} \in \mathfrak{R}^3$ is the bias of acceleration measurements, and $\textbf{b}_{g} \in \mathfrak{R}^3$ is the bias of gyroscope measurements, respectively. }

\subsection{Transition model}\label{subsec:transition_model}

Based on the defined state vector, the state transition is formulated as
\begin{equation}
	\textbf{x}_{k+1} = f\left(\mathbf{x}_k\right)+ \mathbf{{w}}_k=
	\begin{bmatrix}
		f_{pose} \left(\mathbf{x}_{k}^{{[1:23]}},\mathbf{u}_{k}\right)\\
		{f_{bias} \left(\mathbf{x}_{k}^{[11:16]}\right)}
	\end{bmatrix} + {\mathbf{w}}_k,
	\label{equation:modified_VIO:system_model_1}
\end{equation}
where superscript of $\mathbf{x}$ denotes the index{es} of elements in {the} state vector, and the modeling noise ${\mathbf{w}}_k \in \mathfrak{R}^{30}$ is assumed to be the white Gaussian noise {as} ${\mathbf{w}}_k \sim \mathcal{N}({\mathbf{0}},\mathbf{Q})$ {where} $\mathbf{Q} \in \mathfrak{R}^{30 \times 30}$, and $\mathbf{u}_{k} \in \mathfrak{R}^6$ is a control input obtained from the IMU.

The pose state transition model in  (\ref{equation:modified_VIO:system_model_1}) is formulated based on the basic law of {the} uniformly accelerated motion \cite{HOLD:IJRR:2010} {as} 
\begin{equation}
	\begin{array}{ll}
		\mathbf{x}_{k+1}^{{[1:16]}}&=f_{pose} \left(\mathbf{x}_k^{{[1:16]}}\right) 
		= \begin{bmatrix}
			{^W_I\textbf{p}_{1,{k+1}}} \vspace{4pt} \\
			{^W_I\textbf{v}_{{k+1}}} \vspace{4pt} \\
			{^W_I\textbf{q}_{1,{k+1}}}
		\end{bmatrix} \vspace{8pt} \\ 
		& =\begin{bmatrix}
			{^W_I\textbf{p}_{1,{k}}} + {^W_I\textbf{v}_{{k}}}\Delta T + {^W{\textbf{a}_k} \frac{\Delta T^2}{2}} \vspace{4pt} \\
			{^W_I\textbf{v}_{{k}}} + {^W{\textbf{a}_k}\Delta T} \vspace{4pt} \\
			{e}^{- {^I{\textbf{g}_k}} \frac{\Delta T}{2}} \odot {^W_I\textbf{q}_{1,{k}}} 
		\end{bmatrix},
	\end{array}
	\label{equation:modified_VIO:system_model_2}
\end{equation}
where the operator $\odot$ denotes the quaternion product.
To predict {the} next position $ {^W_I\textbf{p}_{{k+1}}}$ and orientation ${^W_I\textbf{q}_{{k+1}}}$ of {an} IMU, 
we use {the} acceleration $^W{\mathbf{a}}$ and {the} angular velocity $^I{\mathbf{w}}$ during {some} time interval. 
They are obtained from {the} inverse process of {generating} inertial measurements \cite{Chatfield:Book:1997} {as} 
\begin{equation}
	\begin{bmatrix}
		{^W\mathbf{a}_k} \\
		{^I{\mathbf{g}}_k} 
	\end{bmatrix}
	{\small = \begin{bmatrix}
		\mathbb{R} \left(^W_I\mathbf{q}_k \right)\left[\mathbf{a}_{m,k}-\mathbf{b}_{a,k} - {\mathbf{w}_{a}}\right] + {^W\mathbf{g}} \vspace{4pt}\\
		{\mathbf{g}}_{m,k} -\mathbf{b}_{g,k} - {\mathbf{w}_{g}} \\
	\end{bmatrix}},
	\label{equation:modified_VIO:system_model_3}
\end{equation} 
where { $\mathbf{{w}}_{a}$ and $\mathbf{{w}}_{g}$ are acceleration and gyroscope measurement noises}, respectively, and assumed to be white Gaussian noise, $\{\mathbf{a}_{m,k},{\mathbf{g}}_{m,k}\} \subset \mathbf{u}_k$ are the measurements from the IMU.
{The} matrix $\mathbb{R}(\cdot)$ denotes a direct cosine matrix converted from the unit-quaternion $\textbf{q}$.
{Then, the second and third poses are transited by the first pose and second poses in the previous time instance. }
\begin{equation}
	\begin{bmatrix}
		{^W_I\textbf{p}_{2,{k+1}}}\\
		{^W_I\textbf{q}_{2,{k+1}}}\\
		{^W_I\textbf{p}_{3,{k+1}}}\\
		{^W_I\textbf{q}_{3,{k+1}}}
	\end{bmatrix}
	=
	\begin{bmatrix}
		{^W_I\textbf{p}_{1,{k}}}\\
		{^W_I\textbf{q}_{1,{k}}}\\
		{^W_I\textbf{p}_{2,{k}}}\\
		{^W_I\textbf{q}_{2,{k}}}
	\end{bmatrix}
\end{equation}

{The bias state transition model in  (\ref{equation:modified_VIO:system_model_1}) is formulated by Brownian motion as $\mathbf{x}_{k+1}^{\left[ 11:16 \right]} = f_{bias}\left( \mathbf{x}_k^{\left[ 11:16 \right]} \right)$.}

\begin{algorithm}[t]
	\caption{{Unscented Kalman Filter (UKF)} with Feature Confidence Analysis} 
	\label{algorithm:modified_VIO:confidence_KF}
	{\small	\begin{spacing}{1.4}
			
		\textbf{$\bullet$ Initialization at $k=0$}
		
		$ \quad \hat{\textbf{x}}_0 =   E[\textbf{x}_{0}], {\textbf{P}_{0}}= E[(\textbf{x}_0-\hat{\textbf{x}}_0)(\textbf{x}_0-\hat{\textbf{x}}_0)^\text{T}] $

		\textbf{$\bullet$ State Estimation with Confidence Analysis, $k\geq 0$}
		
		{\textbf{(1)} Prediction}
		
		$ \quad [\hat{\textbf{x}}_{k+1}^{-},{\textbf{P}_{k+1}^{-}}]=\text{{UKF\_}PREDICT}(\hat{\textbf{x}}_{k}^{+},{\textbf{P}_{k}^{+}},\textbf{Q}_k,f,\textbf{u}_{k})$ 
		
		{\textbf{(2)} 3-points RANSAC with predicted states} 
		
		$ \quad \mathbf{z}_{k+1}^{inlier} = \text{RANSAC}(\mathbf{z}_{k+1},\hat{\textbf{x}}_{k+1}^{-},\theta)$ 
		
		{\textbf{(3)} Confidence inference in Section \ref{sec:infer_confidence} }
		
		$ \quad \mathbf{C}_{z,k+1} = \text{INFER}({\textbf{u}}_{k}, \mathbf{z}_{k+1}^{inlier})$ 	
		
		{\textbf{(4)} Update with confidence in Section \ref{sec:confidence_update}}
		
		$ \quad [\hat{\textbf{x}}_{k+1}^{+},{\textbf{P}_{k+1}^{+}}]	
		=\text{{UKF\_}UPDATE}(\hat{\textbf{x}}_{k+1}^{-},{\textbf{P}_{k+1}^{-}},h,\textbf{z}_{k+1}^{inlier},\mathbf{C}_{z,k+1})$
		
		\vspace{2mm}
		
		$ \rhd$ $f$ is a transition model {in Section \ref{subsec:transition_model}}. 		
		
		$ \rhd$ $h$ is a measurement model {in Section \ref{subsec:measurement_model}}. 
		
		$ \rhd$ $\textbf{u}_{k}$ is a control input, $\textbf{z}_k$ is a measurement.
		
		$ \rhd$ $\theta$ is a threshold for RANSAC. 		
		
	\end{spacing}}
\end{algorithm}

\subsection{Measurement model}\label{subsec:measurement_model}

As measurements for state estimation, {feature points extracted from images are used with trifocal tensor geometry.}
The trifocal tensor encodes the geometric relationship between scene structure and three images that capture the scene at different {viewpoints}.
From a line and the corresponding two points in 3D space and three projection matrices of each image, the trifocal tensor {$\mathbf{T} \in \mathfrak{R}^{3 \times 3 \times 3 }$} is given by
\begin{equation}
	\mathbf{T}_j = \mathbf{P}_{1,4} \left( \mathbf{P}_{2,j} \right)^\top - \mathbf{P}_{2,4} \left( \mathbf{P}_{1,j} \right)^\top,
	\label{equation:modified_VIO:measurement_model_1}
\end{equation}
where subscript $j \in \{1,2,3\}$ denotes {the} column index of a matrix {or a trifocal tensor}, and $\mathbf{P}_i {\in \mathfrak{R}^{3 \times 4}}, \ i \in \{1,2,3\}$, are camera projection matrices of three consecutive frames. For example, $\mathbf{P}_{i, j}$ is the $j$-th column vector of the $i$-th camera projection matrix. {The projection matrices of cameras are acquried from the IMU pose states of  (\ref{equation:modified_VIO:system_model_1}) and transformation parameters $^I_C\mathbf{t},\ ^I_C\mathbf{q}$ between an IMU and a camera. }

A point in {the first} image can be predicted {from} {the} corresponding line in {the} second image and {the corresponding} point in {the third} image with the trifocal tensor.
It is called point-line-point correspondence in literature {given as} 
\begin{equation}
	{\mathbf{f}}_{{1}} = 
	\mathbf{K}\left( \sum_{{j}} ({\mathbf{f}}_{{3,j}}) \mathbf{T}_{{j}}^\top \right)\mathbf{l}_2 ,
	\label{equation:modified_VIO:measurement_model_2}
\end{equation}
where $\mathbf{f}_{{i}} {\in \mathfrak{R^3}}, \mathbf{l}_{{i}} {\in \mathfrak{R^3}}$ {are} a 3D point vector and a line vector in $i$-th camera coordinates, respectively. {$\mathbf{f}_{{i,j}}$ is the {$j$}-th element of {a 3D point vector in the {$i$}-th camera}, and $\mathbf{K}$ is an intrinsic parameter matrix of a camera.}

{A} measurement $\textbf{z}_m \in \mathfrak{R^2}, \ m \in \{1,...,M\}$ in {the first} image is stacked to construct {the} measurement vector $\mathbf{z}$ for motion estimation {as $\mathbf{z} = [\mathbf{z}_1^\top, \cdots, \mathbf{z}_M^\top]^\top$}.
The {measurement} model $h_m$ is {defined} as
\begin{equation}
	\textbf{z}_{m} = 
	\begin{bmatrix}
		u_m\\
		v_m
	\end{bmatrix}
	,
	\label{equation:modified_VIO:measurement_model_3}
\end{equation}
\begin{equation}
	%\mathbf{f}_1^m
	%=
	\begin{bmatrix}
		u_m\\
		v_m\\
		1
	\end{bmatrix}
	=
	h_m(\textbf{x};{\textbf{f}_{3}^m,\textbf{l}_2^m}) + {\mathbf{v}}_m= 
	\mathbf{K}\left( \sum_i ({\mathbf{f}}_{3,i}^m) \mathbf{T}_{{j}}^\top \right)\mathbf{l}_2 + {\mathbf{v}}_m ,
	\label{equation:modified_VIO:measurement_model_3}
\end{equation}
where measurement noise ${\mathbf{v}} \in \mathfrak{R}^{2M}$ is assumed to be white Gaussian noise {as} ${\mathbf{v}} \sim \mathcal{N}({\mathbf{0}},\mathbf{R})$.
%
%The measurement noise covariance matrix $\mathbf{R}$ is defined as $\mathbf{R}= diag\left( \mathbf{R}^1 \cdots \mathbf{R}^M\right)$, where $\mathbf{R}^m=\sigma\mathbf{I}_2$.

%========================================================

\section{Confidence-incorporated Kalman Filter}\label{sec:confidence_analysis}

In this section, we illustrate our main contributions, the analysis of feature confidence in forward motion estimation and the confidence-incorporated Bayesian ego-motion estimation framework.
We {exploit depth cues with the street assumption} and motion cues from inertial measurements for confidence analysis.
{With these cues, our confidence inference algorithm can be} applicable to efficient VIO methods which do not explicitly estimate depth as states.
%
%The inferred feature confidence is imposed into the measurement noise covariance matrix.
%
Algorithm \ref{algorithm:modified_VIO:confidence_KF} shows {the overall framework of} the proposed confidence-incorporated estimation algorithm.
{Unlike \cite{Hu:ICRA:2014}, we use 3-point RANSAC for robustness instead of 1-point RANSAC.}

\subsection{Street Assumption}\label{sec:street_assumption}

We assume the general moving vehicle platform in street environments. 
{Motions} of a vehicle {are} mainly composed of forward and rotation (pitch and yaw) motion{s}.
During forward motion, the static feature points, which are inliers for ego-motion estimation, are extracted from trees, houses (or buildings), and cars along the street.
As a result, the distant feature points are placed in the middle part of the image as shown in Fig. \ref{fig:street_assumption:depth_of_street}.
We statistically analyzed the relationship between the depth of the features and {the} motion axis {for forward motion} by the 16 residential sequences of the public KITTI dataset.
According to our analysis, almost all features are located along the motion axis.
{So, we can conclude that the depth of features and the angles between the motion axis and the 3D position vectors of features are inversely proportional.}
Fig. \ref{fig:street_assumption:distribution} strongly supports our arguments on street assumption.
This shows that the depth of a feature can be roughly inferred from its angle with the motion axis. 
{The geometric definitions of the angle and the depth of a feature and their relation are described in Fig. \ref{fig:confidence_analysis:angle_distance_depth}.}
We exploit this observation to infer the confidence of feature points.
\begin{figure}[t]	
	\centering
	\subfigure[An image that is overlaid with Velodyne depth measurements from residential sequences of the KITTI dataset. The range of the depth is from 0 to 80.  ]
	{\includegraphics[width=0.9\linewidth]{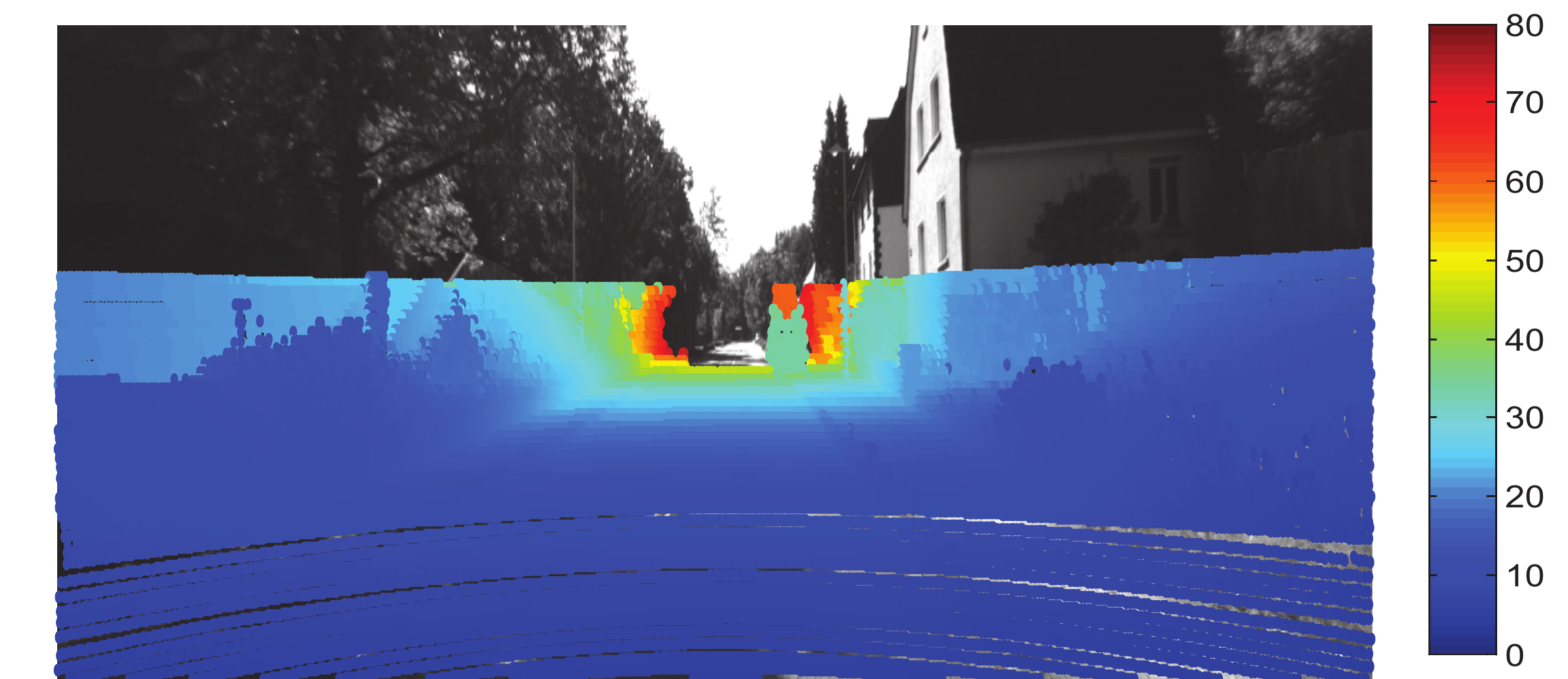}
		\label{fig:street_assumption:depth_of_street} }	
	%	\subfigure[The description on the motion axis vector and a 3D position vector of the feature. When the motion axis are fixed, the angle between both vectors is proportional to the distance between the two vectors in the image plane, and inverse proportional to the depth of the feature under street assumption. ]
	%	{{\includegraphics[width=0.45\linewidth]{./img/street_assumption_9}}
	%	\label{fig:street_assumption:angle}}
	%	\hspace{0.1\baselineskip}
	\subfigure[The 2D histogram of {the depth and the angle of features. The angle is defined with the 3D position vector and the motion axis.} (Total 135,671 features points from the 16 sequences of the KITTI residential dataset are {counted} for the statistics.)]
	{{\includegraphics[width=0.97\linewidth]{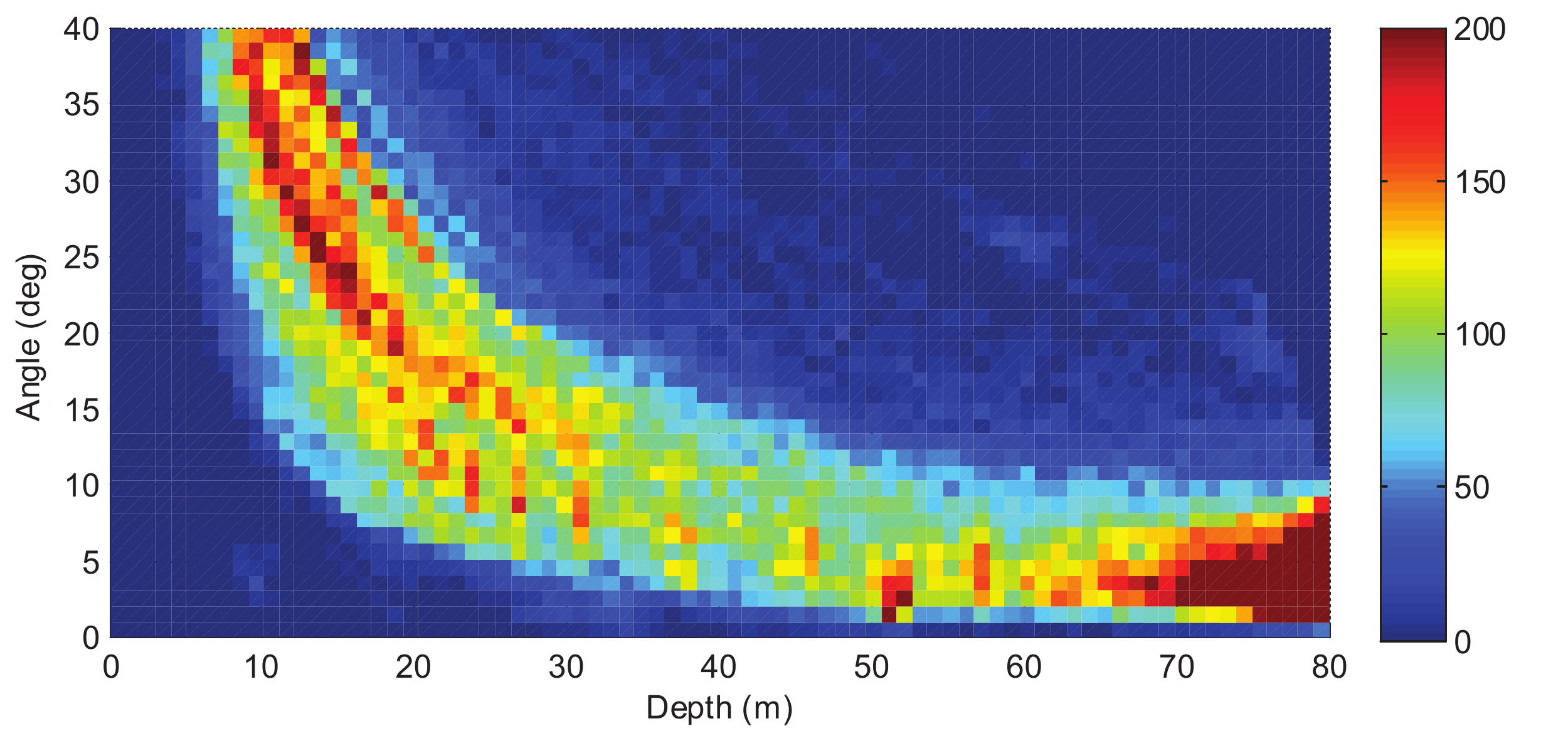}}
		\label{fig:street_assumption:distribution}}
	\caption{The 2D histogram of the depth and the angles of features from the 16 sequences of the KITTI residential dataset} 		
	\label{fig:street_assumption}
\end{figure}
\\

\subsection{Inference of feature confidence}\label{sec:infer_confidence}

{As we illustrated in Section \ref{sec:intro}, distant feature points {located} along the motion direction have low confidence for estimating forward motion.}
{With this intuition, we infer} the confidence of all features in an image. 
We exploit the angle information between the motion axis and the 3D position vector of a feature point mentioned in Sec \ref{sec:street_assumption}.
The angle implicitly includes the depth information in the 3D camera coordinates under the aforementioned street assumption, as well as the {distance information between the motion axis and a feature} in {the} 2D image coordinate {as described in Fig. \ref{fig:confidence_analysis:angle_distance_depth}}.
{For example, if the depth of a feature {increases}, then the angle of the feature {decreases}.}
{If the distance of a feature to the motion axis {decreases}, then the angle of the feature {decreases}.}
{That is, the angle of a feature to the motion axis is proportional to the confidence of a feature for estimating forward motion.}

From this analysis, we define the feature confidence for forward motion estimation based on the motion direction as
\begin{equation}
	c_{t,m} = \text{cos}^{-1} \left( \frac{^C\textbf{v} \cdot {\textbf{f}_1^m}}{\| ^C\textbf{v} \| \| \textbf{f}_1^m \|} \right), 
\end{equation}
where $\left\| \cdot \right\|$ {denotes the magnitude of a vector}, $^C\textbf{f}_1^m {\in \mathfrak{R^3}}$ is {the 3D position vector of} the $m$-th feature in the first camera coordinate among three consecutive camera coordinates in Sec \ref{subsec:measurement_model}, and $^C\mathbf{v}^\top {\in \mathfrak{R^3}}$ is a velocity of {the camera coordinate}.
%
%
%On the other hand, we define the confidence for the rotational motion estimation analogously as
%
%\begin{equation}
%c_{r,m} = \frac{1}{ 1 + {\| ^W_I\textbf{w}^{\top}\|_2} }
%\end{equation}
%
Fig. \ref{fig:confidence_analysis:inertial_confidence} shows feature points with the confidence map inferred by the motion direction for the {forward} motion estimation.

\begin{figure}[t]
	\centering
	\includegraphics[width=0.9\linewidth]{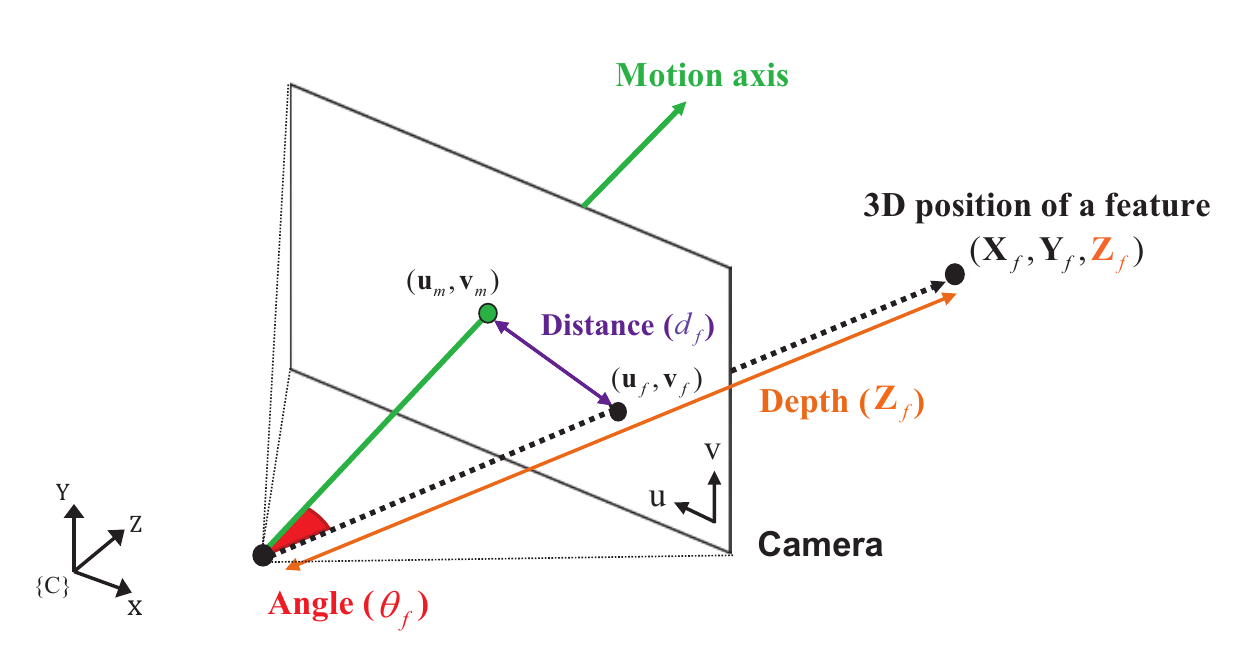}
	\caption{Relation of the motion axis vector and the 3D position vector of a feature. $\theta_f$ is the angle between the motion axis and the 3D position of a feature. 
		$d_f$ is the distance between $\left( u_f, v_f \right)$ and  $\left( u_m, v_m \right)$ in the image coordinate. $Z_f$ is the depth of a feature.}		
	\label{fig:confidence_analysis:angle_distance_depth}
\end{figure}

{However, the matching confidence of feature points is usually opposite to the confidence for estimating forward motion.}
{For example, the matching accuracy of distant features are generally higher than {that} of close features, because the displacements of distant features are small in forward motion.}
{Although it looks like a paradox on feature confidence, we believe that the RANSAC process guarantees the measurement noise of features to be {small and} consistent.}

%When vehicle is moving, the motion is mainly composed of translation. 
%
%{However,} it is difficult to apply above confidences in general {situations} %driving situation.
%
{The inferred feature confidence is only applicable {for} forward motion.}
For this reason, we need to examine {the ratio of forward and rotation motions in a vehicle motion} by inertial measurement analysis.
%
%To handle problems in forward motion, we separate motion to rotation translation by IMU measurement analysis.
%
{We define the} forward motion ratio $\tau$ {as}
\begin{equation}
	\tau = \frac{ \left\| {{^W_I\mathbf{v}}} \right\| }{ \left\| \mathbf{g}_m \right\| +  \left\| ^W_I\mathbf{v} \right\| },
\end{equation}
where $\mathbf{a}_{{m}}$ is acceleration {measurements} and $\mathbf{g}_{{m}}$ is {gyroscope measurements}.
%
% CR: difference of scale
Even though the units of acceleration and angular velocity are different, the motion ratio $\tau$ is empirically {acceptable}, because the magnitudes of two measurements are proportional to the translational and rotational force.
The scale problem caused from different units can be {resolved} by leaning-based regression techniques.
Fig. \ref{fig:confidence_analysis:motion_analysis_imu} shows motion changes {with time and corresponding $\tau$ values obtained} by {the} inertial measurement analysis.

Final confidence of {a featrue is} determined as 
\begin{equation}
	c_{m} = \tau c_{t,m}.
\end{equation}
%
%The ratio plays a key role to impose confidence to feature during forward driving.
%
%However, above defined motion confidence $c_{m}$ is inverse proportional to measurement uncertainty of feature points.

\begin{figure}[t]
	\centering
	\includegraphics[width=0.99\linewidth]{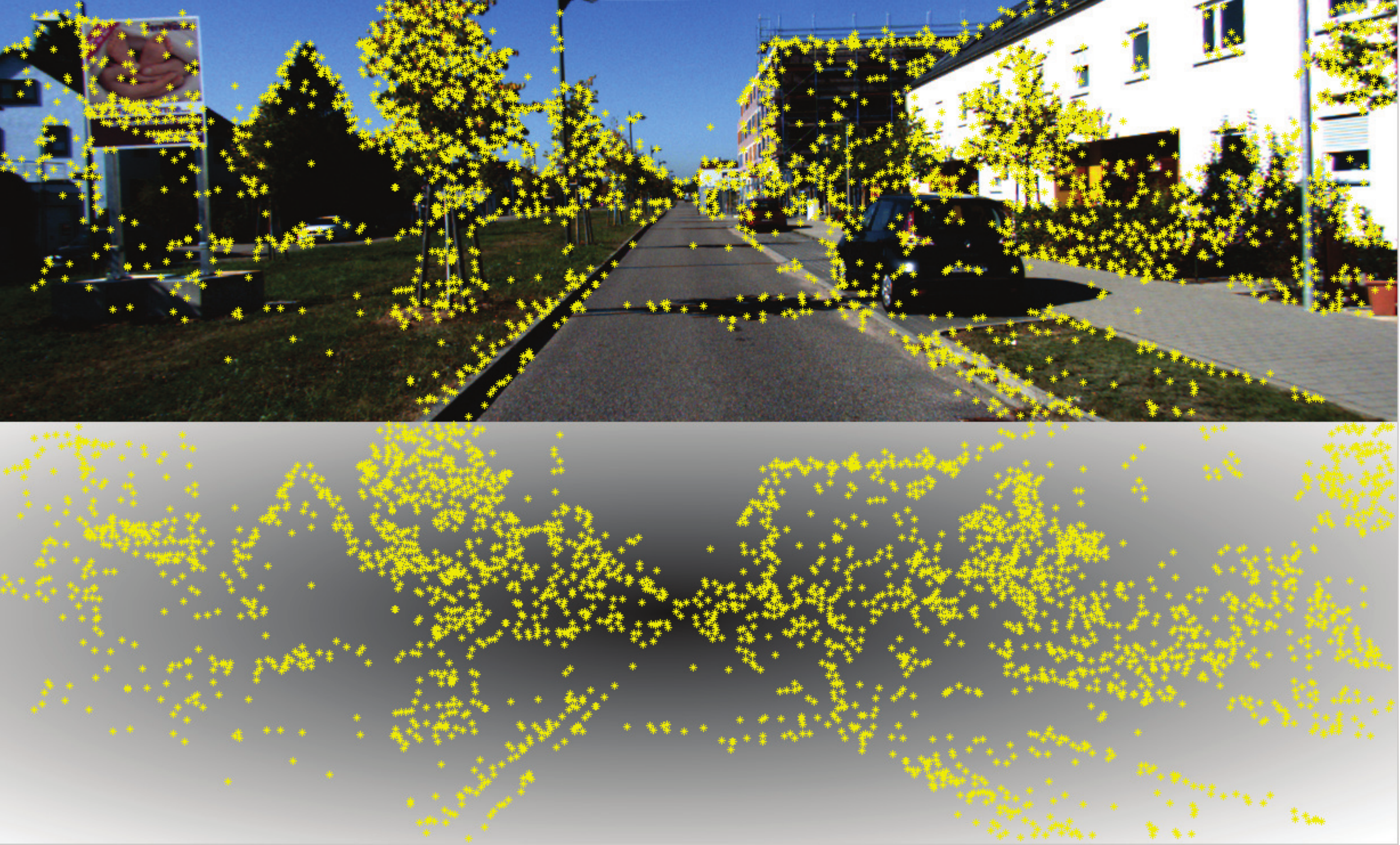}
	\caption{Feature points with {a} confidence map inferred from {the} motion direction ({brighter pixels in the confidence map represent larger confidence}). }		
	\label{fig:confidence_analysis:inertial_confidence}
\end{figure}

\subsection{Update with confidence}\label{sec:confidence_update}

{The Kalman filter {considers} the uncertainties of {a} system and measurements to handle various system {configurations} and measurement noise.}
{Under normal circumstances, the uncertainties are modeled as common white Gaussian noise because it makes {the} system easy to solve.} % and optimal in {the} linear model. }
{ In the case of {image features}, all features are {assumed} to have {the} same uncertainty {in general as}}
\begin{equation}
	\mathbf{z} = 
	h(\mathbf{x}) + \mathbf{{v}},
	\ \ \ p(\mathbf{{v}}) \sim  \mathcal{N}({\mathbf{0}},\mathbf{R}).
	\label{equation:modified_VIO:confidence_update_1}
\end{equation}

{ However, in our framework, we impose different uncertainties to measurement noise covariance $\mathbf{R}$ {based on the feature confidence analysis}.}
{ The {feature} confidences described in {Sec.} \ref{sec:infer_confidence} are block-diagonalized {as}}
\begin{equation}
	\mathbf{C}_{f} = {bl\_diag(c_{1}\mathbf{I}_2,...,c_{M}\mathbf{I}_2}).
	\label{equation:modified_VIO:confidence_update_2}
\end{equation}
Then, the confidence matrix $\mathbf{C}_{f}$ is multiplied to $\mathbf{R}$.
{ As a result, {the} Kalman gain $\mathbf{K}_{c}$ {reflecting the feature confidence} is produced as follows.}
\begin{equation}
	\mathbf{K}_{c} = \mathbf{P}_{xy}(\mathbf{P}_{yy}+\mathbf{C}_{f}\mathbf{R})^{-1}
	\label{equation:modified_VIO:confidence_update_3}
\end{equation}
{Here $\mathbf{P}_{xy}$ is state-measurement cross covariance and $\mathbf{P}_{yy}$ is measurement covariance for UKF Kalman gain.}

{This Kalman gain is used to update state $ \mathbf{x}^{{-}}$ and state error covariance $\mathbf{P}^{{-}}$ {as follows}.}
\begin{equation}
	{\mathbf{\hat{x}}^{+}} = {\mathbf{\hat{x}}^{-}} + \mathbf{K}_{c}(\mathbf{z}-\hat{\mathbf{z}})
	\label{equation:modified_VIO:confidence_update_4}
\end{equation}
\begin{equation}
	{\mathbf{P}^{+}} = {\mathbf{P}^{-}} - \mathbf{K}_{c}\mathbf{P}\mathbf{K}_{c}^\top
	\label{equation:modified_VIO:confidence_update_5}
\end{equation}

\begin{figure}[t]
	\centering
	\includegraphics[width=0.90\linewidth]{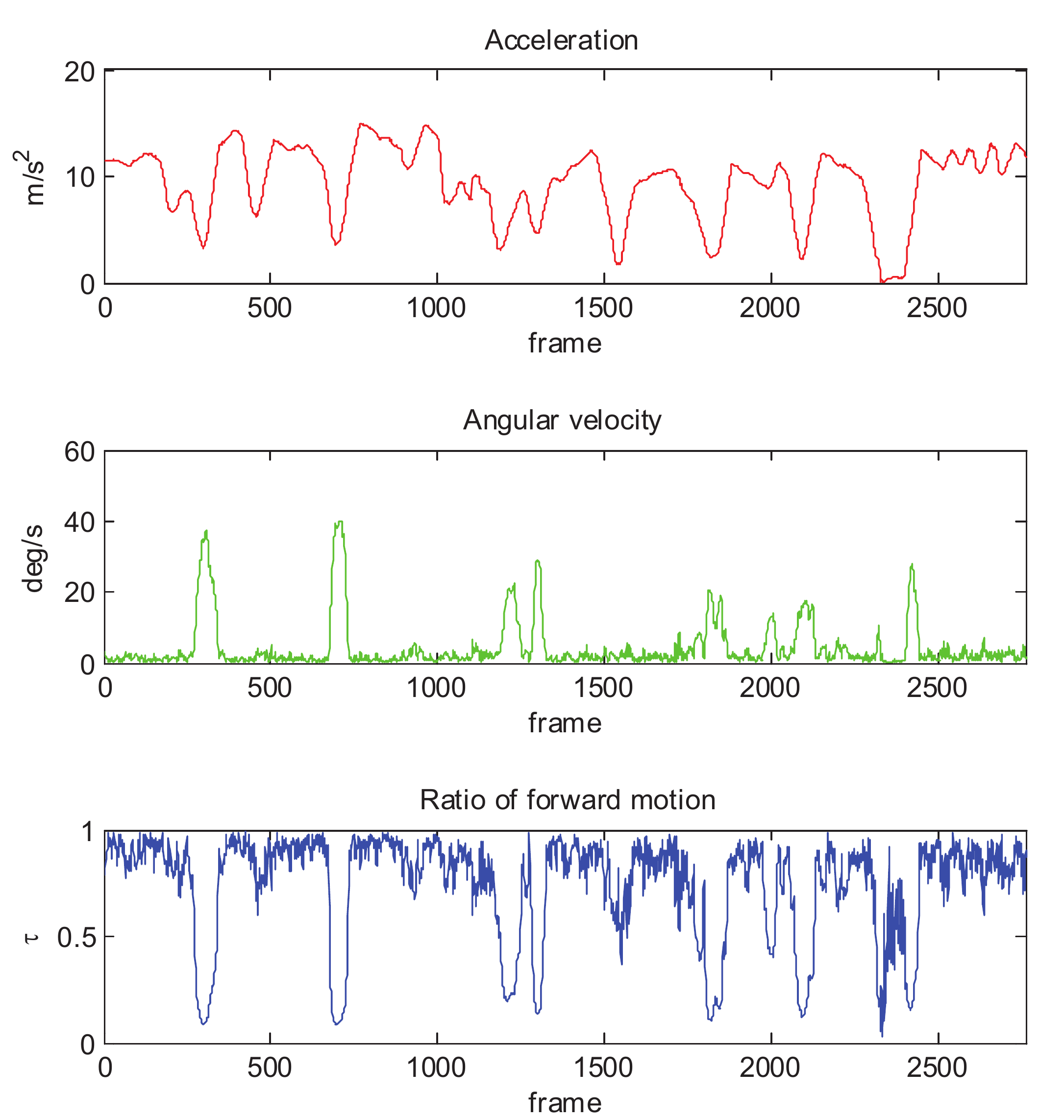}
	\caption{Motion analysis results of the sequence $2011\_09\_30\_drive\_0018$ ($\#10$ in Sec \ref{sec:experimental_results}) {using} inertial measurements  ({\color{red}upper}: acceleration measurements, {\color{green}{middle}}: gyroscope measurements,  {\color{blue}lower}: forward motion ratio). }
	\label{fig:confidence_analysis:motion_analysis_imu}
\end{figure}

\begin{figure}[t]
	\centering
	\subfigure[Original inertial measurements from KITTI dataset]{\includegraphics[width=0.95\linewidth]{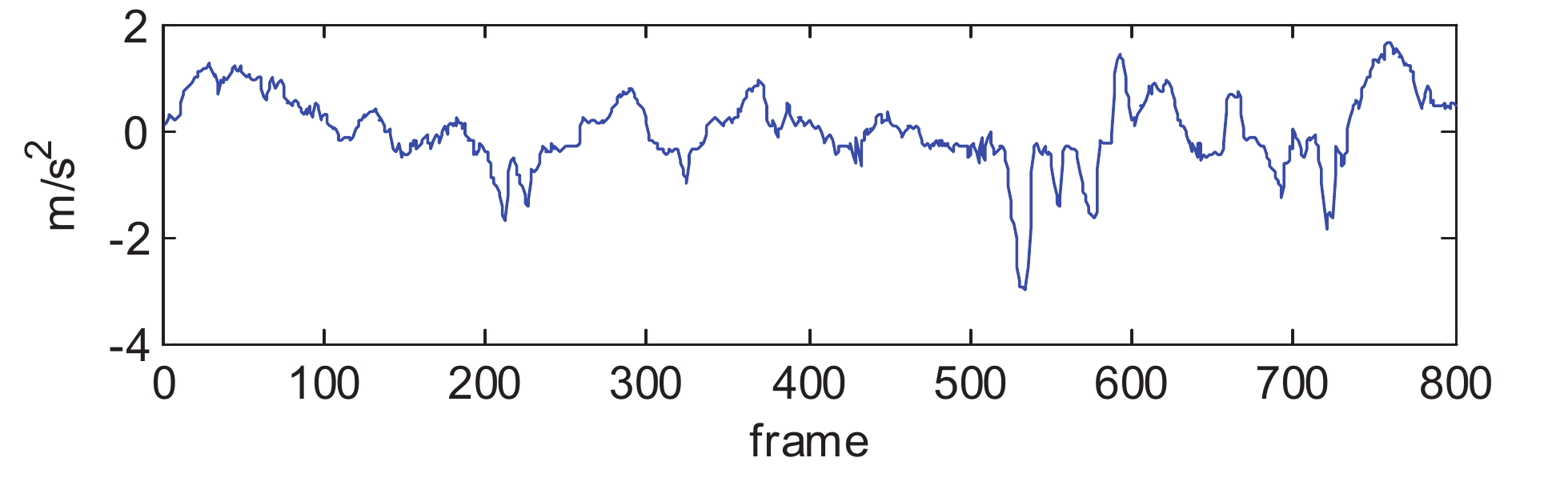}}\\
	\subfigure[Noise-added inertial measurements used in our experiments]{\includegraphics[width=0.95\linewidth]{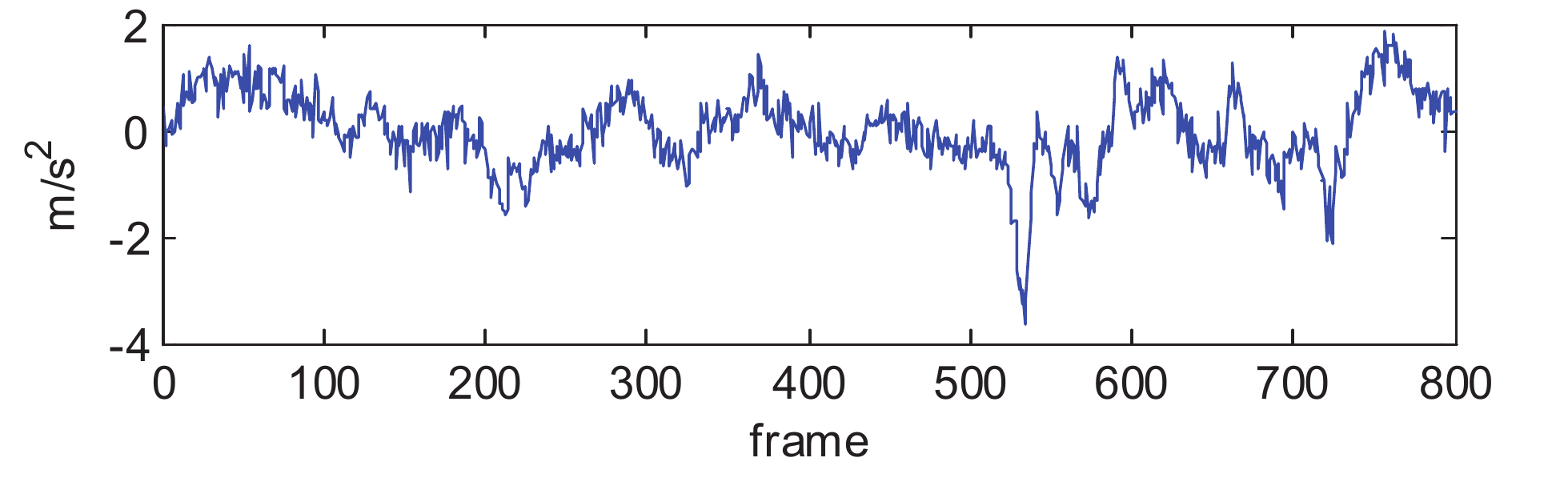}}
	\caption{ Comparison of inertial measurements. Y-axis denotes the acceleration. }
	\label{fig:exp:IMU_measurements}
\end{figure}

\begin{figure*}[t]
	\centering
	\subfigure[ \#6: 2011\_09\_26\_drive\_0061]{\includegraphics[width=0.32\linewidth]{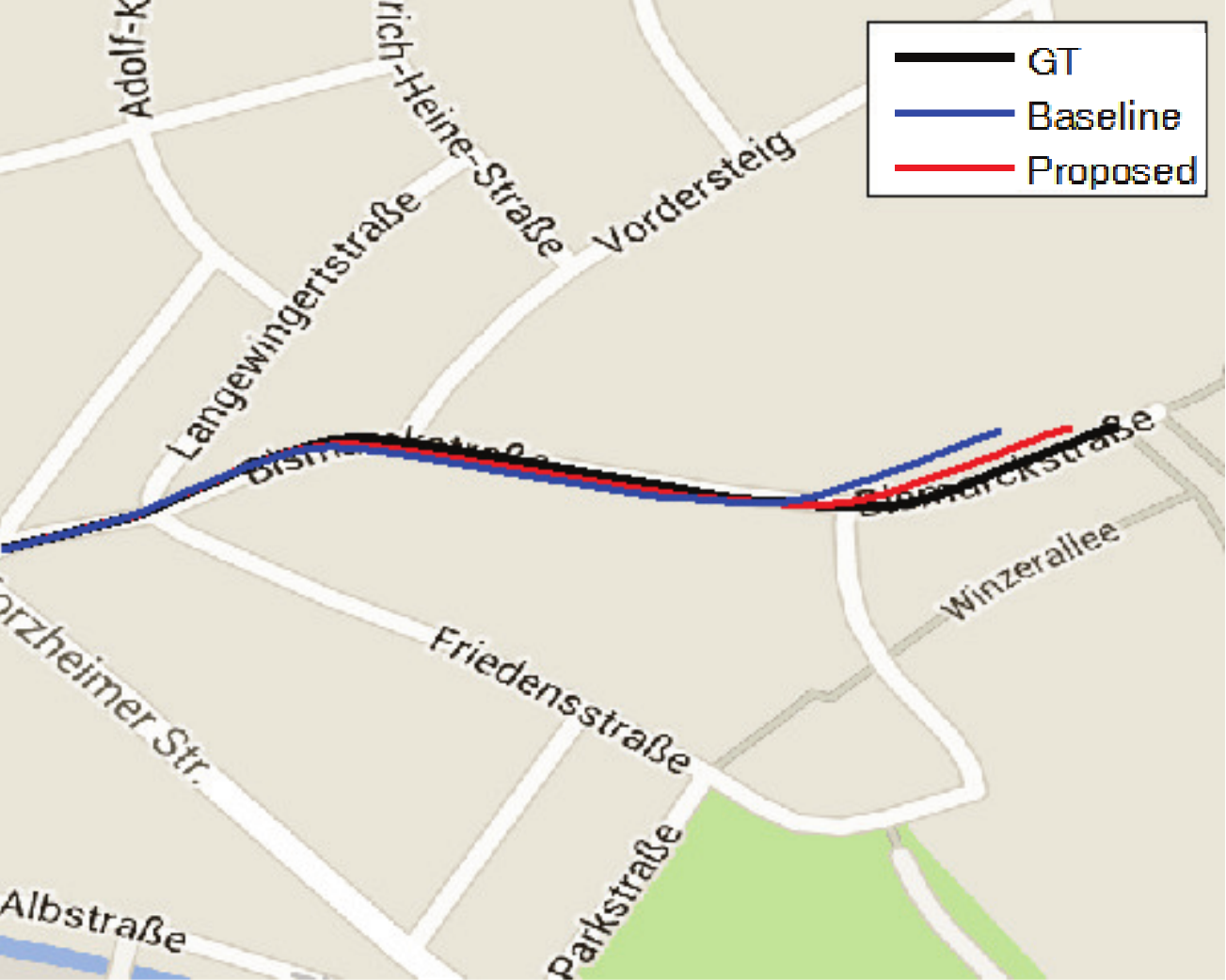}}
	\subfigure[ \#8: 2011\_09\_26\_drive\_0086]{\includegraphics[width=0.32\linewidth]{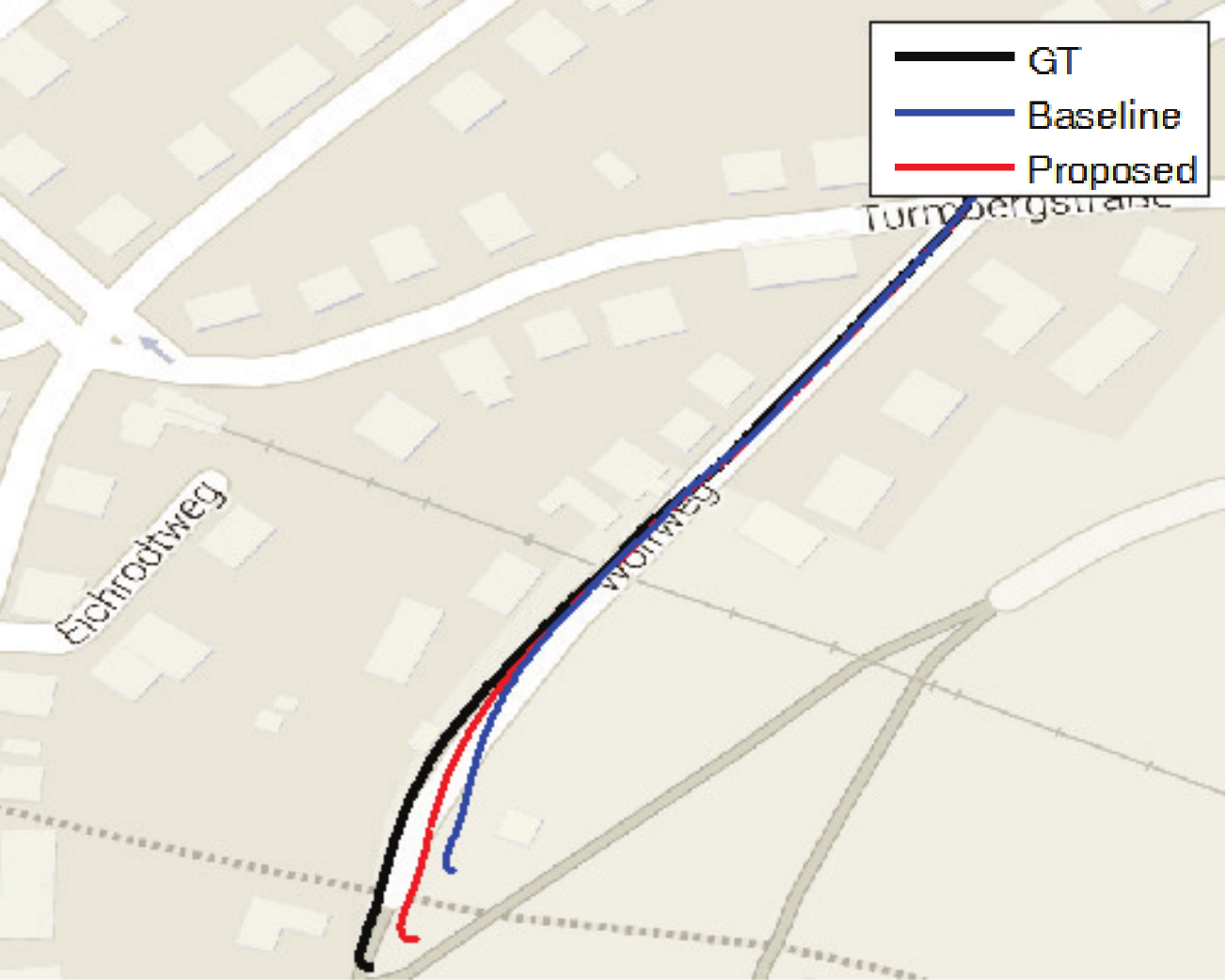}}
	\subfigure[ \#9: 2011\_09\_26\_drive\_0087]{\includegraphics[width=0.32\linewidth]{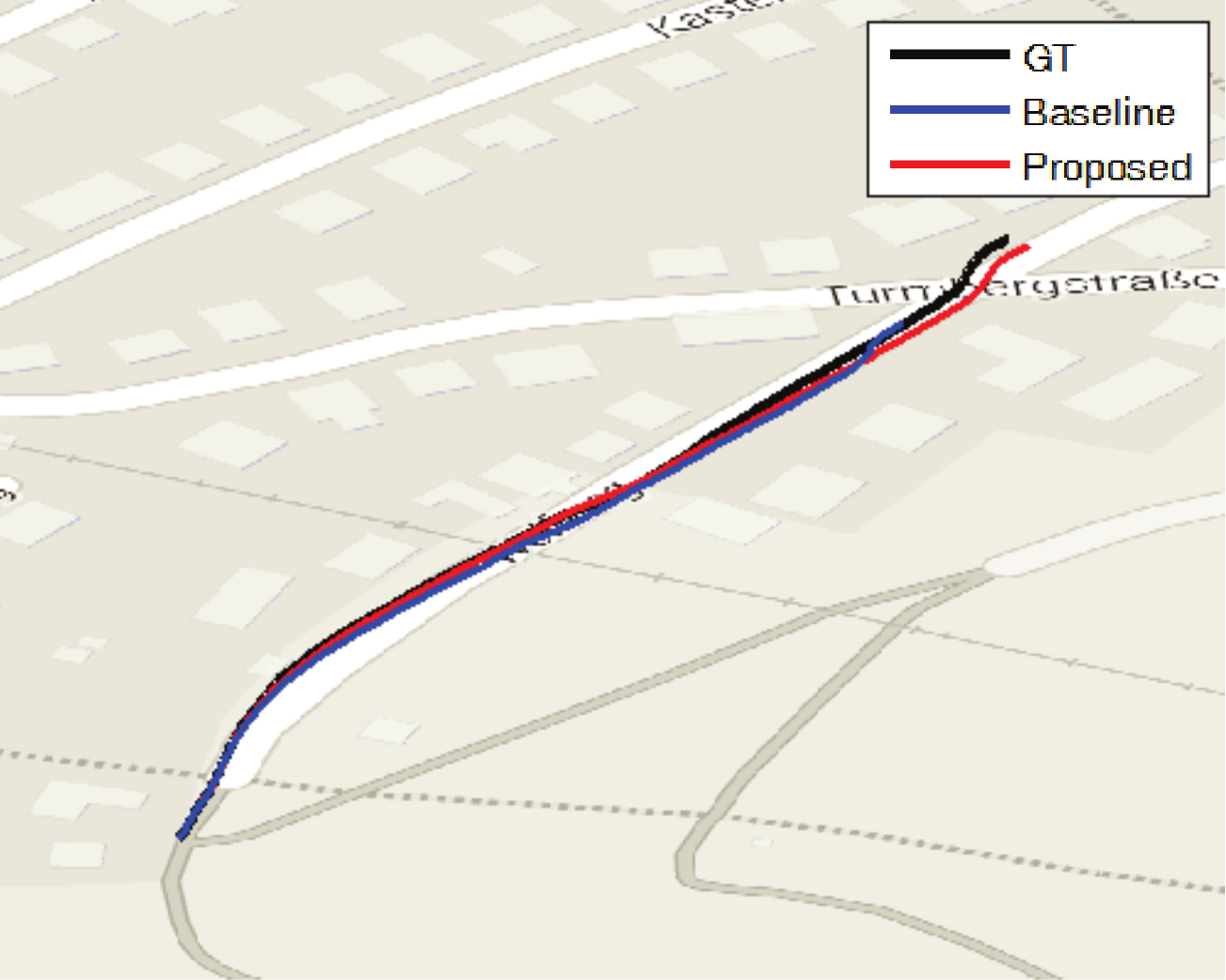}}\\
	\caption{ Selective qualitative comparison results between {the} baseline and {the} proposed algorithm. }
	\label{fig:exp:traj_selection}
\end{figure*}

\section{Experimental Results}\label{sec:experimental_results}

We performed experiments on public KITTI {datasets} \cite{Geiger:CVPR:2012}.
{However, since the} odometry benchmark {datasets} on {the} KITTI {website} do not provide inertial measurements for VIO,  
{we} utilized {KITTI} raw data {datasets} which {include} all information {for VIO}.
We {evaluated} our algorithm on 16 sequences in residential {environments} longer than 100 (m).
Residential sequence\footnote{\url{http://www.cvlibs.net/datasets/kitti/raw_data.php?type=residential}\label{foot:residential}} is the most proper {sequence compared} to City, Road, and Campus sequences because they have small {numbers} of moving {objects}.
%
%Table \ref{table:exp:sequnces} specifies the orders, the names, and {the} lengths of the test sequences {for} our experiments.
%

%
{However,} the inertial measurements {provided from the} KITTI {datasets are} highly accurate {so that they can be} used as ground truth for evaluation.
{On the other hand,} it is well-known that IMU uncertainty is severely high.
We assume general environments that use a less accurate low-cost IMU and camera.
{Therefore,} it is not suitable to use the inertial measurements {from the KITTI dataset directly} as inputs of {the baseline and} the proposed VIO.
{For this reason,} we added Gaussian noise of {the} low-cost IMU( $\sigma_{n_a}$ = 0.25$m/s^2$, $\sigma_{n_g}$ = 0.26$^{\circ}/s$ ) to inertial measurements of KITTI {datasets}.
The noise strength was determined from {the} real data by {using} the technique described in \cite{Woodman:TECH:2007}.
{Fig. \ref{fig:exp:IMU_measurements} shows the original inertial measurements of {the} KITTI dataset and the noisy input used in our experiments.}
{Image features used} for motion estimation {were} extracted and matched by {using the} SIFT descriptor which is {a} well-known robust descriptor.
%
%This can show that our algorithm pushing to high-quality input for motion estimation.
%
For fair evaluation, we adopt evaluation metrics and {codes} from {the} KITTI benchmark \footnote{\url{http://www.cvlibs.net/datasets/kitti/eval_odometry.php}\label{foot:odometry}} .

%\begin{table}[t] 
%	\renewcommand{\tabcolsep}{5mm}
%	\processtable{The 16 sequences of the public KITTI dataset for evaluation. \label{table:exp:sequnces}}
%	{\begin{tabular}{@{\extracolsep{\fill}}lcr} \toprule
%		Sequence  &  Sequence name  &	Length ($\#$ of frames)\\ 	\midrule
%		$\#$1 & 2011\_09\_26\_drive\_0019  & 481 \\ 		
%		$\#$2 & 2011\_09\_26\_drive\_0022  & 800 \\
%		$\#$3 & 2011\_09\_26\_drive\_0023  & 474 \\
%		$\#$4 & 2011\_09\_26\_drive\_0036  & 803 \\			
%		$\#$5 & 2011\_09\_26\_drive\_0039  & 395 \\
%		$\#$6 & 2011\_09\_26\_drive\_0061  & 703 \\
%		$\#$7 & 2011\_09\_26\_drive\_0064  & 570 \\
%		$\#$8 & 2011\_09\_26\_drive\_0086  & 706 \\												
%		$\#$9 & 2011\_09\_26\_drive\_0087  & 729 \\		
%		$\#$10 & 2011\_09\_30\_drive\_0018 & 2380 \\ 		
%		$\#$11 & 2011\_09\_30\_drive\_0020 & 1104 \\		
%		$\#$12 & 2011\_09\_30\_drive\_0027 & 700 \\		
%		$\#$13 & 2011\_09\_30\_drive\_0028 & 5177 \\		
%		$\#$14 & 2011\_09\_30\_drive\_0033 & 1594 \\		
%		$\#$15 & 2011\_09\_30\_drive\_0034 & 1224 \\														
%		$\#$16 & 2011\_10\_03\_drive\_0034 & 4663 \\	\midrule	
%		\multicolumn{2}{c}{Total} & 23291 \\	\botrule		
%	\end{tabular}}{} 	
%\end{table}
 
\begin{table*}[t] 
	\footnotesize
	\centering
	\renewcommand{\tabcolsep}{2.5mm}	
	\caption{Comparison to {the} baseline odometry algorithm on public KITTI benchmark datasets. Rotation and translation errors are evaluated after moving 100 (m) at every frame and are averaged. The numbers in parenthesis indicate the length of the sequence which is the number of frames. }
	\label{table:exp:quantitative_results}
	{\begin{tabular}{@{\extracolsep{\fill}}l|cccc|cccc} \hline 
			\multicolumn{1}{c|}{\multirow{2}{*}{Sequence}} & \multicolumn{4}{c|}{   Rotation error ($10^{-2}$ deg/m) }  & \multicolumn{4}{c}{ Translation error (\%) } \\ \cline{2-9}
			& Baseline \cite{Hu:ICRA:2014}   &   Proposed &  VISO2-M \cite{Geiger:2011:IV} & VISO2-S \cite{Geiger:2011:IV} &   Baseline \cite{Hu:ICRA:2014} & Proposed &  VISO2-M \cite{Geiger:2011:IV} & VISO2-S \cite{Geiger:2011:IV} \\ 	
			\hline 
			$\#$1 (481) & 0.19 &	 0.20 &      1.88  &  0.52     	&  5.16 &  3.99 &       7.23 & 2.33  \\ 		
			$\#$2 (800) & 0.86 &    0.86 &       4.89  &  1.81        & 2.58 &	 3.37 &      15.16 & 2.89	  \\
			$\#$3 (474)& 0.29 &	 0.30 &        1.76   &   0.82     	 &	4.20 &	 3.83 &      19.81  & 2.65		\\	
			$\#$4 (803) & 0.34  &	 	0.32 &       2.79 &   1.57      	 &	3.42 &	 	2.61 &       11.09  & 	1.67	\\		
			$\#$5 (395) & 0.15 &	 	0.17  &       0.66 &  0.62      	 &	6.64 &	 	4.21 &       15.52 & 1.38   \\
			$\#$6 (703) & 0.31 &	 	0.30  &       1.32 &  1.34        	 &	11.67 &	 	5.38 &       49.70 & 1.97  			 \\
			$\#$7 (570) & 0.27  &	 	0.29 &       3.29 &  1.01       &	6.67 &	 	7.61 &      133.26  &1.52  			  \\
			$\#$8 (706) & 0.40  &  	0.41  &        2.59 &  1.69        &	13.10 &	  	4.90 &      236.60 & 2.29  	 \\
			$\#$9 (729) & 0.77  &  	0.78 &        2.70 &   1.15       	  &	13.19 &	  	3.99 &      294.81 & 2.63   	 		 \\
			$\#$10 (2380) & 0.46  &	  	0.46 &       5.52 &  1.84       &	5.00 &	  	4.29 &      71.83  & 2.00   \\												
			$\#$11 (1104) & 0.32  & 	0.32 &       3.11 & 1.47       &   	4.60 &  	4.10 &       39.13  & 1.97			   \\		
			$\#$12 (700) & 0.42  &	  	0.41 &       7.77 & 2.49       &	  	1.69 &	  	2.02 &       62.91  & 2.21 	 		 \\ 		
			$\#$13 (5177) & 0.54  &	  	0.53 &       3.33 & 1.86        &	  5.94 &	  	5.56 &      53.75   & 3.94 	 		  \\		
			$\#$14 (1594) & 0.38  &  	0.37 &       2.62 &  1.50       &	  	4.86 & 	4.40  &       4.55 & 1.84  		 \\		
			$\#$15 (1224) & 0.41  &  	0.41 &       4.72 &  1.44       &	  	4.76 & 	5.24  &       37.34 & 1.22  	  \\		
			$\#$16 (4663) & 0.46 &  	0.46 &       1.98 &  1.20      &	  	12.79 &  	9.76  &       9.58 & 1.48  		  \\	\hline 		 
			All (23291) & 0.45 &	  	0.45 &       3.37 &    1.55    &	  	7.19	& 		5.63  &        52.38 & 2.33 	\\ 	\hline	
			\label{table:exp:baseline_quantitative_results}	
		\end{tabular}}{} 	
	\end{table*}

We exploited the Hu's work as the baseline odometry \cite{Hu:ICRA:2014}. 
However, the proposed confidence analysis can be inco{r}porated into any {Bayesian estimation framework} as we already mentioned in Section \ref{sec:intro}.
All parameters for baseline and proposed VIO such as initial error covariance $\mathbf{P}$, system noise covariance $\mathbf{Q}$, measurement noise covariance $\mathbf{R}$, and RANSAC threshold are equally set {for} fair evaluation.
{Besides,} input feature points are also identical in all evaluation{s}. 
The maximum number of feature points are set to 100 with bucketing.
Estimates of rotation and translation are averaged in 10-Monte Carlo simulation.
Table {\ref{table:exp:baseline_quantitative_results}} shows that {the} our simple confidence incorporation leads to {the} quantitative performance {improvement} of VIO.
The improvement appears in translation estimation as we intended.
%
%Fig. \ref{fig:exp:translation_error} visually expresses the translation errors of Table {\ref{table:exp:baseline_quantitative_results}}.
%
We can notice that the proposed algorithm really suppresses the translation error in sequence $\#$ 6, 8, 9.
The qualitative comparison of these sequences are shown in Fig. \ref{fig:exp:traj_selection}.
The rotation estimates of the proposed algorithm have negligible variation compare to {the} baseline.
The accuracy of rotation estimates really high thanks the gyroscope measurements from the IMU though.

\section{Discussion and Conclusion} \label{sec:conclusion}

In this paper, we proposed the robust VIO algorithm based on the analysis of feature confidence in ego-motion estimation using inertial measurements. We first formulate the VIO problem by using effective trifocal tensor geometry. Then, by using the motion information obtained from inertial measurements, we define the feature confidence in ego-motion estimation and propose the confidence-incorporated ego-motion estimation framework based on inertial measurements. 
Experimental results on the public KITTI dataset show that the proposed algorithm outperforms the baseline VIO, and it demonstrates the effectiveness of the proposed feature confidence analysis and the proposed confidence-incorporated ego-motion estimation framework. 

We expect that the proposed algorithm can be applied to the smart-phone platform equipped with a low-cost IMU and a camera.
Furthermore, thanks to the advances of various computer vision techniques, some other cues such as depth inferred from single image can be incorporated into our framework to extend our work.

\bibliographystyle{IEEEtranS}
\bibliography{feature_analysis}

\end{document}